\documentclass{new_tlp}
\usepackage{mathptmx}
\usepackage{amsmath}
\usepackage{graphicx}
\usepackage{multirow}
\usepackage{epstopdf}
\usepackage{lastpage}
\usepackage{amssymb}

\newtheorem{example}{Example}
\newtheorem{theorem}{Theorem}

\newtheorem{corollary}{Corollary}
\newtheorem{proposition}{Proposition}

\newcommand{\system}[1]{\textsc{#1}}

\newcommand{\isect}{\cap}
\newcommand{\Isect}{\bigcap}
\newcommand{\union}{\cup}
\newcommand{\Union}{\bigcup}

\newcommand{\pair}[2]{\langle{#1},{#2}\rangle}
\newcommand{\tuple}[1]{\langle{#1}\rangle}
\newcommand{\set}[1]{\{#1\}}
\newcommand{\sel}[2]{\{{#1}\mid{#2}\}}
\newcommand{\rg}[3]{{#1}\,{#2}\ldots{#2}\,{#3}}
\newcommand{\eset}[2]{\set{\rg{#1}{,}{#2}}}

\newcommand{\IF}{\leftarrow}
\newcommand{\AND}{,\,}
\newcommand{\naf}{\mathtt{not}\,\,}
\newcommand{\END}{.}

\newcommand{\hd}[1]{\mathrm{head}(#1)}
\newcommand{\bd}[1]{\mathrm{body}(#1)}
\newcommand{\pbd}[1]{\mathrm{body}^+(#1)}
\newcommand{\nbd}[1]{\mathrm{body}^-(#1)}

\newcommand{\sig}[1]{\mathrm{At}(#1)}
\newcommand{\lm}[1]{\mathrm{LM}(#1)}
\newcommand{\GLred}[2]{{#1}^{#2}}

\newcommand{\at}[1]{\mathtt{#1}}
\newcommand{\trop}[1]{\mathrm{Tr_{#1}}}
\newcommand{\tr}[2]{\trop{#1}(#2)}
\newcommand{\suppr}[2]{\mathrm{SR}_{#1}(#2)}
\newcommand{\pdep}{\mathbin{\succeq}}
\newcommand{\dgplus}[1]{\mathrm{DG}^+(#1)}

\newcommand\bcmdtab{\noindent\bgroup\tabcolsep=0pt%
  \begin{tabular}{@{}p{10pc}@{}p{20pc}@{}}}
\newcommand\ecmdtab{\end{tabular}\egroup}

  \title[Theory and Practice of Logic Programming]
        {Capturing (Optimal) Relaxed Plans with  Stable and Supported Models of Logic Programs}

  \author[M. F. Rankooh and T. Janhunen]
         {MASOOD FEYZBAKHSH RANKOOH and TOMI JANHUNEN\\
         Tampere University, Tampere, Finland\\
         \email{masood.feyzbakhshrankooh@tuni.fi, tomi.janhunen@tuni.fi}}

\jdate{March 2003}
\pubyear{2003}
\pagerange{\pageref{firstpage}--\pageref{lastpage}}
\doi{S1471068401001193}

\begin{document}

\label{firstpage}

\maketitle

  \begin{abstract}
We establish a novel relation between delete-free planning, an important task for the AI Planning community also known as relaxed planning, and logic programming. We show that given a planning problem, all subsets of actions that could be ordered to produce relaxed plans for the problem can be bijectively captured with stable models of a logic program describing the corresponding relaxed planning problem. We also consider the supported model semantics of logic programs, and introduce one causal and one diagnostic encoding of the relaxed planning problem as logic programs, both capturing relaxed plans with their supported models. Our experimental results show that these new encodings can provide major performance gain when computing optimal relaxed plans, with our diagnostic encoding outperforming state-of-the-art approaches to relaxed planning regardless of the given time limit when measured on a wide collection of STRIPS planning benchmarks.
  \end{abstract}

  \begin{keywords}
    Planning heuristics, Answer Set Programming, Optimal delete-free planning, Supported models, Acyclicity constraint, Dependency graph, Vertex elimination
  \end{keywords}

\tableofcontents

\section{Introduction}
AI Planning, an active research area of Artificial Intelligence, is the task of finding a sequence of actions, called a plan, that when applied to a given initial state transforms it to a state that satisfies all members of a given set of goal conditions. According to the STRIPS formulation of AI Planning, states and goal conditions are represented by sets of atomic propositions, and each action can have separate sets of atomic propositions as its preconditions, positive effects (also called add effects), and negative effects (also called delete effects). Delete-free planning problems are those for which actions have no negative effects. A given Planning problem can be relaxed into a delete-free problem, optimal solving of which provides lower bound of the optimal cost of the original problem. This lower bound, denoted by $h^+$, could be used as a heuristic in an A*-like search scheme to find an optimal solution for the original problem. Computing $h^+$ is, however, NP-equivalent \cite{DBLP:journals/ai/Bylander94}. Also, $h^+$ is hard to approximate \cite{DBLP:conf/ki/BetzH09}.

Optimally solving relaxed planning problems in an efficient way is important for multiple reasons. There have been many admissible heuristic functions that approximate $h^+$ in polynomial time by computing lower bounds. Examples are $h^{max}$ heuristic \cite{DBLP:journals/ai/BonetG01}, LM-cut heuristic \cite{DBLP:conf/aips/HelmertD09}, set-additive heuristic \cite{DBLP:conf/ecai/KeyderG08}, and cost-sharing approximations of $h^{max}$ \cite{DBLP:conf/aips/MirkisD07}. The informativeness of these heuristic functions cannot be evaluated unless we can compute the exact value of $h^+$. Using such a measure for informativeness could lead to devising more informative heuristic functions. Moreover, efficient solving of relaxed planning problems is in itself of importance, because there exist planning tasks of interest for the AI community whose actions are all delete-free. Examples of such tasks are the minimal seed-set problem \cite{DBLP:conf/aips/GefenB11}, and the problem of determining join orders in relational database query plan generation \cite{DBLP:conf/aaai/RobinsonMT14}. Another reason for the importance of efficient optimal relaxed planning is the fact that optimal plans for non-relaxed planning problems can always be produced by iterative solving and reformulating relaxed planning tasks \cite{DBLP:conf/aips/Haslum12}. By repeatedly finding optimal plans for newly produced relaxed problems, while reformulating the non-relaxed problem in each iteration, one can reach a point where the found optimal plan for the last relaxed problem is actually an optimal plan for the original problem.

Several approaches to solving relaxed planning problems have previously been introduced. The approaches include Boolean satisfiability (SAT) based encodings \cite{RankoohRintanen22DR}, integer programming based models \cite{DBLP:journals/jair/ImaiF15,RankoohRintanen22ICAPS}; and a minimum-cost hitting set based method introduced by \citeN{DBLP:conf/aips/HaslumST12}. In this work we take a new approach based on the stable and supported models of logic programs \cite{GL88:iclp,MS92:tcs}.
%
Such models provide the semantical basis for answer set programming
(ASP); see, e.g., \cite{BET11:cacm} for an overview.  The ASP
paradigm offers general-purpose modeling languages for knowledge representation
and reasoning.

A typical encoding of a search problem in ASP aims at a
one-to-one correspondence between answer sets and the solutions of the
problem. This is in perfect harmony with AI planning where sequences
of actions (plans) form solutions to problems at hand.  Indeed, many
AI planning problems have been encoded as logic programs \cite{SB18:ki}
and AI planning also played a role in the early development of the ASP
paradigm \cite{Lifschitz99:iclp} in the first place.
Both stable and supported models implement a form of minimality, i.e.,
atomic propositions are \emph{false by default}. This is highly useful
in the context of AI planning since state predicates are falsified in
this sense and the encodings of planning problems can concentrate on
specifying which state predicates become true or remain true
inertially.  This tends to lead to more compact encodings compared to
those based on pure SAT and, furthermore, enable memory savings if native
answer set solvers are used for actual computations.
The difference between stable and supported models is also interesting
in this setting, since ASP solvers may compute answer sets based on
either semantics. Stable models are also supported models but not vice
versa in general. The gap between the two semantics vanishes if a
logic program is suitably instrumented, e.g., in terms of acyclicity
constraints \cite{BGJKS16:fi}. These observations open up new avenues
when it comes to encoding planning problems as logic programs as well
as choosing an approach for computing plans as answer sets.

In this work, we establish a new relation between relaxed planning and logic programs. We give an encoding of relaxed planning problems in ASP. We show that all subsets of actions that could be ordered to produce relaxed plans can be bijectively captured with stable models of the produced logic program. This enables the previously uninvestigated usage of off-the-shelf answer set solvers for computing the value of $h^+$. While the supported model semantics of logic programs cannot be directly employed for this purpose, we show how by guaranteeing acyclicity in an underlying graph of the logic program, one may deploy supported models to harvest (optimal) relaxed plans of the planning problem. The logic program produced in this way inherits the causal nature of our stable model based encoding, in the sense that the direction of explanations provided by the rules is from causes/preconditions to effects. By reversing this direction, we provide a diagnostic encoding, which while still using the supported model semantics of logic programs, is shown to be more efficient than our causal encoding by our empirical study. Our experimental results show that when given small time limits these new encodings can significantly outperform the previous  approaches to relaxed planning when measured on STRIPS planning benchmarks. Moreover, regardless of the used time limit, our diagnostic supported model based encoding enables \system{Clasp} \cite{GKK0S15:lpnmr} to solve more problems compared to the integer programming solver based state-of-the-art method.

Logic programming has recently been employed for computing heuristics for lifted planning tasks. \citeANP{DBLP:conf/aaai/CorreaPHF22} \citeyear{DBLP:conf/aips/CorreaFPH21,DBLP:conf/aaai/CorreaPHF22} employed Datalog programs to calculate $h^{add}$ \cite{DBLP:journals/ai/BonetG01} and $h^{FF}$ \cite{HoffmannN01}, respectively, for lifted planning tasks. However, the objective of our work differs from theirs. While both $h^{add}$ and $h^{FF}$ are non-admissible estimations of $h^+$ and can be computed in polynomial time for ground instances, we aim to compute $h^+$ itself. Furthermore, this work focuses on ground planning tasks. Although the generalization of our current approach to lifted planning is relatively simple, we leave it for future research.

The rest of this article is organized as follows. In Section \ref{section:preliminaries}, we recall basic concepts and definitions of planning problems, relaxed planning, logic programs, and their stable and supported model semantics. Then, in Section \ref{section:stable}, we show how relaxed plans can be captured with stable models of an encoding of relaxed planning problems into logic programs. In Section \ref{section:stable}, we first show how a logic program can be augmented with a dynamically varying digraph whose acyclicity guarantees a shift in the semantics from stable models to supported models. We then  recall how vertex elimination can be used to check whether a given digraph is acyclic. Based on the supported model semantics and the vertex elimination method, we explain our causal and diagnostic encodings of relaxed planning problems. We present practical evidence in Section \ref{section:experiments} based on an experimental evaluation of the resulting encoding for answer set and supported model optimization. This analysis is based on 2212 problem instances from 84 STRIPS planning problem sets. Finally, we conclude the paper in Section \ref{section:conclusion}.

\section{Preliminaries}
\label{section:preliminaries}
Since we intend to establish a connection between AI Planning and Answer Set Programming,  we provide necessary formal definitions with respect to both of these paradigms.

\subsection{AI Planning and relaxed plans}

A STRIPS planning problem is a 5-tuple $\Pi =  \tuple{X, I, A, G,cost}$ where $X$ is a finite set of Boolean state variables, also called \emph{atomic propositions}. The initial state $I$ and the set of goal conditions $G$, are subsets of $X$. The finite set $A$ is the set of actions. Each member $\vec{a}$ of $A$ is a triple $\tuple{pre(\vec{a}),add(\vec{a}),del(\vec{a})}$, where $pre(\vec{a})$, $add(\vec{a})$ and $del(\vec{a})$ are sets of atomic propositions denoting the set of preconditions, positive effects, and negative effects of $\vec{a}$, which are the atomic propositions that $\vec{a}$ requires, adds, and deletes, respectively. The cost function $cost$ maps members of $A$ to a non-negative integer. We use the vector sign to distinguish actions from the corresponding atoms that represent them in logic programs.

States are represented as subsets of $X$. The successor $s' = exec_{\vec{a}}(s)$ of a state $s$ with respect to action $\vec{a}\in A$ is defined if $pre(\vec{a})\subseteq s$, where the definition is $s'=(s\setminus del(\vec{a}))\cup add(\vec{a})$. An action sequence $\vec{a_1}, ... ,\vec{a_n}$ is executable (in state $s$) if $exec_{\vec{a_1},...,\vec{a_n}}(s)=exec_{\vec{a_n}}(...exec_{\vec{a_2}}(exec_{\vec{a_1}}(s))...)$ is defined. A plan for $\Pi$ is a sequence $\pi$ of actions from $A$ such that $G\subseteq exec_\pi(I)$. The cost of plan $\pi= \vec{a_1},...,\vec{a_n}$ for $\Pi$, is defined by $\Sigma_{i=1,...,n} cost(\vec{a_i})$. An optimal plan for $\Pi$ is a plan with minimal cost.

For a given STRIPS planning problem $\Pi =  \tuple{X, I, A, G,cost}$, the delete relaxation~\cite{DBLP:journals/ai/BonetG01} is defined as $\Pi^+ =  \tuple{X, I, A^+, G,cost}$, where $A^+$ is defined from $A$ by replacing the set of negative effects of each member of $A$ with the empty set. Without loss of generality, we can define $\Pi^+ =  \tuple{X, \emptyset, A^+, G,cost}$, with an additional requirement that all members of $I$ have been removed from $G$, and also from the preconditions and effects of members of $A^+$. We use this latter definition of  relaxation in the rest of the paper.

A plan for $\Pi^+$ is called a relaxed plan for the original problem $\Pi$. The minimal cost of plans of $\Pi^+$ is denoted by $h^+(\Pi)$. If there is no relaxed plan for $\Pi$, we set $h^+(\Pi)$ to $\infty$.

\subsection{Answer set programming}

In this work, we consider logic programs that
consist of \emph{rules} of the forms:
\begin{align}
\label{eq:normal-rule}
& \phantom{\set{}}
a\IF\rg{b_1}{\AND}{b_n} \AND \rg{\naf c_1}{\AND}{\naf c_m}\END
\\
\label{eq:choice-rule}
&
\set{a}\IF\rg{b_1}{\AND}{b_n} \AND \rg{\naf c_1}{\AND}{\naf c_m}\END
\end{align}
The symbols $a$, $\rg{b_1}{,}{b_n}$ with $n\ge 0$, and
$\rg{c_1}{,}{c_m}$ with $m\ge 0$ occurring in the rules are
(propositional) \emph{atoms} and ``$\naf\!\!$'' denotes
\emph{negation by default}.
Rules of the forms \eqref{eq:normal-rule} and \eqref{eq:choice-rule} are
known as \emph{normal} and \emph{choice} \emph{rules},
respectively \cite{SNS02:aij}. Intuitively, each rule $r$ gives a
reason to derive its \emph{head} $\hd{r}=a$ if the conditions in its
\emph{body} $\bd{r}$ are met, i.e., atoms involved can be either
derived or not by other rules. For a choice rule $r$ of form
\eqref{eq:choice-rule}, the derivation of $\hd{r}$ is optional, enabling an exception to $\hd{r}$ being false by default.
We write $\pbd{r}$ and $\nbd{r}$ for the sets of atoms
$\rg{b_1}{,}{b_n}$ (resp.~$\rg{c_1}{,}{c_m}$) occurring positively
(resp.~negatively) in $\bd{r}$. We say that $r$ is a positive rule if $\nbd{r}$ is empty.

The \emph{signature} of a logic program $P$ is the set of atoms
$\sig{P}=\Union_{r\in P}(\set{\hd{r}}\union\pbd{r}\union\nbd{r})$
that occur in $P$.
The \emph{positive dependency graph} of $P$ is
$\dgplus{P}=\tuple{\sig{P},\pdep}$ where $a\pdep b$ holds for
$a,b\in\sig{P}$ if $\hd{r}=a$ and $b\in\pbd{r}$ for some rule $r\in
P$. If $a\pdep b$, we say that $a$ depends on $b$, and also denote this by $\pair{a}{b}\in\dgplus{P}$.

An \emph{interpretation} $I\subseteq\sig{P}$ determines which atoms
$a\in\sig{P}$ are \emph{true} ($a\in I$) and which are \emph{false}
($a\not\in I$).
Then $I$ satisfies a rule $r\in P$ of form \eqref{eq:normal-rule}, denoted $I\models r$, if the satisfaction of
the body, denoted $I\models\bd{r}$, implies that $\hd{r}\in I$, i.e.,
$I\models\hd{r}$. For a choice rule $r$ of form
\eqref{eq:choice-rule}, $I\models r$ unconditionally. Moreover, the
interpretation $I$ is a (classical) \emph{model} of $P$ if $I\models
r$ holds for every $r\in P$.
Each positive normal program $P$ has a unique \emph{least model}
$\lm{P}$ obtained as the intersection
$\Isect\sel{I\subseteq\sig{P}}{I\models P}$.

Given an interpretation $I$, the \emph{reduct} $\GLred{r}{I}$ of $r$
with respect to $I$ is obtained by partially evaluating the negative
conditions of $r$. For a normal rule \eqref{eq:normal-rule},
$\GLred{r}{I}=\emptyset$ if $c_i\in I$ for some $1\leq i\leq m$ and
$\GLred{r}{I}=\set{a\IF\rg{b_1}{\AND}{b_n}}$ otherwise. For a choice
rule \eqref{eq:choice-rule}, the latter case additionally requires
that $a\in I$. 
Finally, for an entire logic program $P$, the reduct
$\GLred{P}{I}=\Union_{r\in P}{\GLred{r}{I}}$ and $I$ is a
\emph{stable model} of $P$ iff $I=\lm{\GLred{P}{I}}$.
For the purposes of this work, it is also useful to distinguish
the \emph{supporting rules} of $P$ with respect to $I$, denoted by 
$\suppr{P}{I}$, which are the normal rules whose bodies are satisfied, and the choice rules whose bodies and heads are satisfied.
Then, a model $I\models P$ is \emph{supported} (by $P$) when
$I=\sel{\hd{r}}{r\in\suppr{P}{I}}$. Each stable model of $P$
is supported, but supported models are not necessarily stable,
such as $I=\set{a}$ for $P=\set{a\IF a\END}$.

\section{Relaxed plans captured with stable models of logic programs}
\label{section:stable}
Typically, modeling planning problems as answer set programs is done by assuming a number of time steps for the output plan, which is also mirrored in the structure of the produced logic program \cite{SonBTM06}. Here, however, we show that, as long as finding relaxed plans are concerned, one can encode the planning problem in such a way that there will be no need for a multi-step structure. 

Let $\Pi =  \tuple{X, I, A, G,cost}$ be a relaxed STRIPS planning problem, $\Pi^+ =  \tuple{X, \emptyset, A^+, G,cost}$ be the delete relaxation of $\Pi$, and $P$ be a logic program consisting of rules of the form (1) $g\IF\naf g$ for every $g\in G$; (2) $\set{a}\IF\rg{q_1}{\AND}{q_n}$ for every $\vec{a}\in A$ with $pre(\vec{a})=\set{\rg{q_1}{\AND}{q_n}}$; (3) $p\IF a$ for every $\vec{a}\in A$ and $p\in add(\vec{a})$. Intuitively, the first rule guarantees all goal atoms to be true in a model. The second rule explains the necessary conditions for the execution of an action $\vec{a}$. The third rule enforces the positive effects in case $\vec{a}$ has been chosen to be in the model.

We show that more relaxed semantics of models could not play the same role. Example 1 shows that neither the classical models nor the supported models of $P$ are generally suitable for capturing the relaxed plans of $\Pi$ correctly. 

\begin{example}
Consider a planning problem $\Pi =  \tuple{X, I, A, G,cost}$,  where $X=\set{p,q}$, $I=\emptyset$, $G=\set{p}$,  $A=\set {\vec{a},\vec{b}}$, $pre(\vec{a})=add(\vec{b})=\set{p}$, $add(\vec{a})=pre(\vec{b})=\set{q}$, and the cost function $cost$ is arbitrary. This problem has no relaxed plan, as $\vec{a}$ and $\vec{b}$ are codependent. The logic program $P$ explained above consists of the following rules:

\begin{center}
$\begin{array}{l@{\hspace{2em}}l}
\set{a}\IF p\END &
\set{b}\IF q\END \\
q\IF a\END &
p\IF b\END \\

p \IF \naf p\END \\

\end{array}$
\end{center}
It is easy to check that $M=\set{a,b,p,q}$ is both a classical and a supported model for $P$. However, $P$ has no stable model, due to circularities involved in the encoding. \hfill $\blacksquare$
\end{example}

\noindent We now formally show that $P$ captures the relaxed plans of $\Pi$ as its stable models.

\begin{theorem} \label{theorem:st-plan} There is a bijection $f(A') = \bigcup_{\vec{a}\in A'} ( add(\vec{a}) \cup \set{a})$ between all subsets $A'$ of $A^+$ which can be ordered to produce a relaxed plan for $\Pi$, and all stable models of $P$.
\end{theorem}
\begin{proof}

We first show that $f$ is well-defined, i.e., if $\pi= \vec{a}_1,...,\vec{a}_m$ is a permutation of members of $A'$ such that $\pi$ is a relaxed plan for $\Pi$, then $M=f(A')$ is a stable model of $P$. For every $g\in G$, $g$ must be added by some action in $\pi$. Thus, the reduct $\GLred{P}{M}$ consists of rules of the form (1) $a\IF\rg{q_1}{\AND}{q_n}$ for every $\vec{a}\in \pi$ and $pre(\vec{a})=\set{\rg{q_1}{\AND}{q_n}}$, and (2) $p\IF a$ for every $\vec{a}\in \pi$ and $p\in add(\vec{a})$. Clearly, $M$ is model for $\GLred{P}{M}$. By bounded induction on the lengths of  prefixes of $\pi$, we show that $M$ is a subset of any model for $\GLred{P}{M}$. As we explained above, the initial state of the relaxed problem is (safely) assumed to be an empty set. Therefore, $\vec{a}_1$ cannot have any precondition. Thus, $\GLred{P}{M}$ includes a rule of the form ($a_1.$), and $ add(\vec{a}_1) \cup \set{a_1}$ is a subset of any model for $\GLred{P}{M}$.
Assume that for $1\le j < m$, $\bigcup_{i=1,...,j} add(\vec{a}_i) \cup \set{a_1,...,a_j}$  is a subset of any model for $\GLred{P}{M}$. Since $\vec{a}_{j+1}$ is executable in $exec_{\vec{a}_1,...,\vec{a}_j}(\emptyset)$, $pre(\vec{a}_{j+1})$ is a subset of $\bigcup_{i=1,...,j} add(\vec{a}_i)$. Because $\GLred{P}{M}$ includes the two types of rules explained above for $\vec{a}_{j+1}$, we conclude that $\bigcup_{i=1,...,j+1} (add(\vec{a}_i) \cup \set{a_i})$  is a subset of any model for $\GLred{P}{M}$.\\

Clearly, $f$ is injective. We now show that $f$ is also surjective, i.e., if $M$ is a stable model of $P$, then there exists $A'\subseteq A^+$ such that $M=f(A')$, and $A'$ can be permuted to produce a relaxed plan for $\Pi$. Let $A'=\{\vec{a} \mid a\in M\}$. We have $G\subseteq M$ because for every $g\in G$, $P$ includes the rule $g\IF\naf g$. The reduct $\GLred{P}{M}$ consists of rules of the form (1) $a\IF\rg{q_1}{\AND}{q_n}$ for every $\vec{a}\in A'$ and $pre(\vec{a})=\set{\rg{q_1}{\AND}{q_n}}$ and (2) $p\IF a$ for every $\vec{a}\in A'$ and $p\in add(\vec{a})$. If $p$ is added by some action $\vec{a}\in A'$, then clearly we must have $p\in M$. On the other hand, for every $p\in X$ if $p\in M$, then $p$ is added by some action $\vec{a}\in A'$, otherwise $M\setminus \set{p}$ would also be a model for $\GLred{P}{M}$, contradicting that $M$ is the least model for $\GLred{P}{M}$. We conclude that $M=f(A')$ and if $A'$ can be ordered to produce a sequence of actions executable in $I$, then that sequence is also a relaxed plan for $\Pi$. 

For the sake of contradiction, assume that $A'$ cannot be ordered to produce a sequence of actions executable in $I$. Let $A''$ be a (possibly empty) proper subset of $A'$ such that its members (if any) can be ordered to produce a sequence of actions executable in $I$, and furthermore, let $A''$ be maximal in the sense that there is no subset of $A'$ with such a property that is also a proper superset of $A''$. Let $M'=\bigcup_{\vec{a}\in A''} add(\vec{a}) \cup \set{a\mid\vec{a}\in A''}$. Clearly, $M'$ is a proper subset of $M$. Let $\vec{a}\in A'$ and $pre(\vec{a})=\set{\rg{q_1}{\AND}{q_n}}$. If $\vec{a}\in A''$, $M'$ trivially satisfies $a\IF\rg{q_1}{\AND}{q_n}$. On the other hand, for every $\vec{a}\in A' \setminus A''$, the maximality of $A''$ implies that at least one precondition of $\vec{a}$ is not in $M'$, and therefore, $a\IF\rg{q_1}{\AND}{q_n}$ is vacuously satisfied. We conclude that $M'$ is a model for $\GLred{P}{M}$, contradicting that $M$ is the least model for $\GLred{P}{M}$. \hfill
\end{proof}

Theorem 1 shows that if $P$ is augmented with an optimization constraint requiring minimization over the summation of the costs of actions in the answer sets, the cost of an optimal stable model of $P$ is equal to $h^+(\Pi)$.

The program $P$ can be seen as a \emph{causal} encoding of relaxed plans of $P$. That is because the direction of explaining the logic of relaxed plan computation in $P$ is from preconditions to actions, and from actions to effects. In other words, the direction is from causes to effects. Alternatively, a \emph{diagnostic} encoding would explain the logic of relaxed plan computation from effects to actions, and from actions to preconditions. In the next section, we show how this latter paradigm could be used for computing relaxed plans.

\section{Relaxed plans captured with supported models of logic programs}
\label{section:supported}
In this section, we recall the instrumentation of logic programs with acyclicity constraint, which allows capturing the stable models of a given logic program $P$ with the supported models of another program $\tr{ACYC}{P}$ which are acyclic with respect to an underlying graph \cite{BGJKS16:fi}. We provide an adaptation of this method based on the structure of program $P$ explained above. We then review the so-called vertex elimination method, used previously for cycle prevention in the produced models of SAT formulas \cite{RR22,DBLP:conf/lpnmr/RankoohJ22}. We next show how vertex elimination could also be used to translate $\tr{ACYC}{P}$ to a new program $P_c$ such that the supported models of $P_c$ represent acyclic supported models of  $\tr{ACYC}{P}$, and thus,  stable models of $P$ and relaxed plans of $\Pi$. Based on the structure of $P_c$, we introduce another logic program $P_d$ which describes the relaxed plans diagnostically. We prove that the supported models of $P_d$ represent those of $P_c$, thereby capturing the stable models of $P$ and relaxed plans of $\Pi$. 
\subsection{Instrumentation of logic programs with acyclicity constraint}

We adopt the \emph{acyclicity translation}
$\tr{ACYC}{P}$ of a logic program $P$ \cite{BGJKS16:fi} that deploys special
\emph{dependency atoms} $\at{dep}(x,y)$
to express the activation of the respective arc
$\pair{x}{y}\in\dgplus{P}$ in the acyclicity constraint. For the sake of the compactness of the output program, instead of using the exact method, we customize the translation method considering the structure of the program $P$ explained above. In particular, we circumvent the introduction of dependency atoms for actions, by establishing dependencies only between atoms of the original planning problem. This way, the underlying graphs for which acyclicity must be guaranteed become considerably smaller than $\dgplus{P}$.

The idea is to \emph{instrument} $P$ explained in the previous section with additional rules that capture
\emph{well-support} for atoms $p\in X$. For each pair
$\tuple{p,q}$, if there exists $\vec{a}\in A$ such that $p\in add(a)$ and $q\in pre(a)$, the potential dependency of $p$ on $q$ is expressed using a choice rule
$\set{\at{dep}(p,q)}\IF q$.
Also, atoms $\rg{\at{ws}(a_1,p)}{,}{\at{ws}(a_k,p)}$, for actions  $\eset{\vec{a}_1}{\vec{a}_k}$ that add $p$ enforce the
well-support for $p$ in terms of $k$ rules
$p \IF\at{ws}(a_i,p)$
for $i=1,...,k$.
For an atom $p\in X$,  the rule \eqref{eq:well-support} below captures the option that the well-support for $p$ is provided by some action $\vec{a}$ such that $pre(\vec{a})=\set{\rg{q_1}{\AND}{q_n}}$ and $p\in add(\vec{a})$.
\begin{align}
\label{eq:well-support}
\set{\at{ws}(a,p)}\IF  \rg{\at{dep}(p,q_1)}{\AND}{\at{dep}(p,q_n)}\END
\end{align}
Also, the rule $a \IF\at{ws}(a,p)$ captures the atom $a$ in the supported models, in the case that it has been used to provide well-support for $p$. As in program $P$, we need a rule $g\IF\naf g$ for every $g\in G$ to guarantee that every goal atom has been produced. 

For $\tr{ACYC}{P}$ obtained in this way, the distinction
between stable and supported models disappears if we insist on
\emph{acyclic models} $I$ for which the digraph induced by the set of
arcs $\sel{\pair{a}{b}}{\at{dep}(a,b)\in I}$ is acyclic. We deploy the following result:

\begin{proposition}[\protect{\citeN{BGJKS16:fi}}]
\label{prop:acyc}
If $M$ is a stable model of $P$, then $\tr{ACYC}{P}$ has an acyclic
supported model $N$ such that $M=N\isect\sig{P}$. If $N$ is an acyclic supported model of $\tr{ACYC}{P}$, then $M=N\isect\sig{P}$ is a stable model of $P$.

\end{proposition}

\begin{example}
Consider $\Pi$ to be the planning problem of Example 1. The program $\tr{ACYC}{P}$ consists of the following rules:

\begin{center}
$\begin{array}{l@{\hspace{2em}}l}
\set{\at{dep}(p,q)}\IF q\END &
\set{\at{dep}(q,p)}\IF p\END \\
\set{\at{ws}(a,q)}\IF\at{dep}(q,p)\END &
\set{\at{ws}(b,p)}\IF\at{dep}(p,q)\END \\
%
q \IF \at{ws}(a,q)\END &
p \IF \at{ws}(b,p)\END \\
a \IF \at{ws}(a,q)\END &
b \IF \at{ws}(b,p)\END \\
p \IF \naf p\END \\

\end{array}$
\end{center}
 It can easily be checked that $M=\set{a,b,p,q,\at{ws}(a,q),\at{ws}(b,p),\at{dep}(p,q),\at{dep}(q,p)}$ is the only supported model for $\tr{ACYC}{P}$. However, this model is not acyclic, as it contains both $\at{dep}(p,q)$ and $\at{dep}(q,p)$. \hfill $\blacksquare$
\end{example}

Similarly to the stable model based encoding, $\tr{ACYC}{P}$ is a causal encoding, expressing the inference in the direction from preconditions to actions, and from actions to effects. However, there are additional concepts in this encoding, namely dependencies and well-support. In fact, in $\tr{ACYC}{P}$, preconditions are assumed to cause dependencies, which in turn cause well-support and effects. Here, well-support atoms $\at{ws}(a,p)$ take the causal role that action atoms $a$ have in $P$. The action atoms are only included in $\tr{ACYC}{P}$ to represent their cost in the minimization constraint. The rules in Example 2 establish the inference direction from preconditions to dependencies (the first row), from dependencies to well-support (the second row), and from well-support to effects (the third and the fourth rows). The final rule captures the goal condition (as before).

\subsection{Vertex elimination graphs}
\label{section:vertex-elimination}

The concept of vertex elimination graphs has been recently shown effective for guaranteeing acyclicity in constraint programs with underlying graphs. The concept of vertex elimination for digraphs was originally introduced by \citeN{ve:ve}.

Given a digraph $ \mathcal{G}=\pair{V}{E}$, an ordering of V is a bijection
$\alpha:\{1,\ldots,n\}\to V$.
For a vertex $v$, the \emph{fill-in} of $v$, denoted by $F(v)$,
is the set of arcs from the in-neighbors of $v$ to the out-neighbors of $v$,
formally defined by 
\begin{equation}
F(v) = \{ \pair{x}{y} \mid \pair{x}{v}\in E, \pair{v}{y}\in E, x\ne y\}.\label{fillinv}
\end{equation}
The $v$-\emph{elimination} graph of $\mathcal{G}$ is produced by removing $v$ from $\mathcal{G}$, and adding the fill-in of $v$ to the resulting graph. Formally, $\mathcal{G}(v)=\pair{V\setminus\{v\}}{E(v)\cup F(v)}$, where $E(v)=\{\pair{x}{y}\mid\pair{x}{y} \in E, x\ne v, y\ne v\}$.

Given a digraph $\mathcal{G}$ and an ordering $\alpha$ of its vertices, the \emph{elimination process} of $\mathcal{G}$ according to $\alpha$ is the sequence $\mathcal{G}= \mathcal{G}_0,  \mathcal{G}_{1},\ldots, \mathcal{G}_{n-1}$, where $\mathcal{G}_{i}$ is the $\alpha(i)$-elimination graph of $\mathcal{G}_{i-1}$ for $i=1,\ldots,n-1$.

The fill-in of the digraph $\mathcal{G}$ according to $\alpha$, denoted by $F_\alpha( \mathcal{G})$, is the set of all arcs added to $\mathcal{G}$ in the vertex elimination process. Formally, $F_\alpha( \mathcal{G})$ is defined by (\ref{filling}), where $F_{i-1}(\alpha(i))$ is the fill-in of $\alpha(i)$ in $\mathcal{G}_{i-1}$:
\begin{equation}
F_\alpha( \mathcal{G})=\bigcup\limits_{i=1}^{|V|-1} F_{i-1}(\alpha(i)).\label{filling}
\end{equation}
The vertex elimination graph of $\mathcal{G}$ according to $\alpha$, denoted by $\mathcal{G}_\alpha^*$, is the union of all graphs produced in the elimination process of $\mathcal{G}$ according to $\alpha$:
\begin{equation}
 \mathcal{G}_\alpha^*=\pair{V}{E\cup F_\alpha( \mathcal{G})}.
\end{equation}

For any digraph $\mathcal{G}$, the number of arcs of the vertex elimination graph depends on the ordering function $\alpha$. It has been shown that the problem of finding the optimal ordering function, the one resulting in the smallest number of arcs in the vertex elimination graph, is NP-complete~\cite{ve:ve}. Nevertheless, there are effective heuristics for finding empirically useful orderings. Examples are the \emph{minimum fill-in} and  \emph{minimum degree} that accordingly choose a vertex for removal at each step during the elimination process. One important property of vertex elimination is that if the original graph $\mathcal{G}$ has a directed cycle, then $\mathcal{G}_\alpha^*$ will have a cycle of length 2, regardless of the ordering $\alpha$. 

\subsection{The causal encoding based on supported models}

Consider $\tr{ACYC}{P}$ explained above. Let  $\mathcal{G}$ be the graph of all dependencies of $\tr{ACYC}{P}$. Formally, $\mathcal{G}=\pair{X}{E}$, where $E=\set{ \pair{p}{q}\mid\at{dep}(p,q)\in \sig{\tr{ACYC}{P}}}$. Also, for each supported model $M$ of $\tr{ACYC}{P}$, let $\mathcal{G}_M$ be the graph of all dependencies in $M$, i.e.,  $\mathcal{G}_M=\pair{X}{E_M}$, where $E_M=\set{ \pair{p}{q}\mid\at{dep}(p,q)\in M}$. Assume that $\alpha$ is an ordering of the members of $X$,  $\mathcal{G}= \mathcal{G}_0, \mathcal{G}_{1},\ldots, \mathcal{G}_{n-1}$ is the elimination process of $\mathcal{G}$ according to $\alpha$, and for $i=1,\ldots,n$, $F_{i-1}(\alpha(i))$ is the fill-in of $\alpha(i)$ in $\mathcal{G}_{i-1}$. Let  $\mathcal{G}_\alpha^*=\pair{X}{E^*}$ and $\mathcal{G}_{M,\alpha}^*=\pair{X}{E_M^*}$ be the vertex elimination graphs of $\mathcal{G}$ and $\mathcal{G}_M$ according to $\alpha$, respectively.

We produce the causal supported model semantics based encoding of $\Pi$ as logic program $P_c$ by adding the following rules to $\tr{ACYC}{P}$. For every $\pair{p}{q}\in F_{i-1}(\alpha(i))$, add
\begin{equation}
\label{rule:acyc1}
\at{dep}(p,q)\IF\at{dep}(p,\alpha(i))\AND \at{dep}(\alpha(i),q).
\end{equation}

\noindent Also, for every $p$ and $q$ such that $\pair{p}{q}\in  \mathcal{G}_\alpha^*$ and $\pair{q}{p}\in  \mathcal{G}_\alpha^*$, we add
\begin{equation}
\label{rule:acyc2}
f\IF\at{dep}(p,q)\AND \at{dep}(q,p)\AND \naf f.
\end{equation}

Intuitively, for any vertex ordering $\alpha$, and any supported model $M$ of $\tr{ACYC}{P}$, the rule (\ref{rule:acyc1}) extends $M$ by atoms representing the arcs in $\mathcal{G}_{M,\alpha}^*$, the vertex elimination graph of $\mathcal{G}_M$ according to $\alpha$, while the rule (\ref{rule:acyc2}) guarantees that $\mathcal{G}_{M,\alpha}^*$ has no cycle of length 2.

\begin{theorem}
\label{theorem:causal}
Let $A'$ be a subset of $A^+$. There exists a permutation $\pi$ of members of $A'$ such that $\pi$ is a relaxed plan for $\Pi$ iff $P_c$ has a supported model $M$ such that $A' = \set{\vec{a}\mid a\in M}$.
\end{theorem}

\begin{proof}
($\implies$) If $\Pi$ has a relaxed plan $\pi= \vec{a}_1,...,\vec{a}_m$, then according to Theorem 1, $\bigcup_{i=1,...,m} add(\vec{a}_i) \cup \set{a_1,...,a_m}$ is a stable model of $P$. By Proposition \ref{prop:acyc}, $\tr{ACYC}{P}$ has an acyclic
supported model $N$ such that $\{ a\in N \mid \vec{a}\in A^+ \} = \set{a_1,...,a_m}$. Let $\mathcal{G}_N=\pair{X}{E_N}$, where $E_N=\set{ \pair{p}{q}\mid\at{dep}(p,q)\in N}$, and let $\mathcal{G}_{N,\alpha}^*=\pair{X}{E_N^*}$ be the vertex elimination graph of $\mathcal{G}_N$ according to $\alpha$. Since $\mathcal{G}_N$ is acyclic, $X$ can be ordered by topological sorting according to $\mathcal{G}_N$. Now, if the vertex elimination process adds the arc $\pair{p}{q}$, then $p$ must be ordered before $q$ by the topological sorting. Therefore, $\mathcal{G}_{N,\alpha}^*$ is also acyclic. It should now be easy to check that $N\cup \set{\at{dep}(p,q)\mid\pair{p}{q}\in E_N^*}$ is a supported model of $P_c$.\\
($\impliedby$) Let $M$ be a supported model for $P_c$. We first show that $M$ is acyclic. Let $\mathcal{G}_M=\pair{X}{E_M}$, where $E_M=\set{ \pair{p}{q}\mid\at{dep}(p,q)\in M}$. Assume that $k>1$ is the smallest number for which there exist a cycle of length $k$ in $\mathcal{G}_M$. Then there are atoms $\at{dep}(p_1,p_2),...,\at{dep}(p_{k-1},p_k),\at{dep}(p_{k},p_1)$ in $M$.  According to the rule (\ref{rule:acyc2}), $k$ cannot be equal to 2. Let $i=argmin_{1\le j\le k} \alpha^{-1}(p_j)$. Then $p_i$ is the vertex in the mentioned cycle that is eliminated before all other vertices in the cycle according to $\alpha$. According to the rule (\ref{rule:acyc1}), $\at{dep}(p_{i-1},p_{i+1})\in M$ (with indices considered modulo $k$), and therefore $\mathcal{G}_M$ has a cycle of length $k-1$, a contradiction. Let $N=M\isect\sig{ \tr{ACYC}{P}}$. A straightforward investigation shows that $N$ is a supported model of $\tr{ACYC}{P}$. By Proposition 1, $N' = N\isect\sig{P}$ is a stable model of $P$. Since $A'=\set{ \vec{a}\in A^+ \mid a\in N'}$, by Theorem 1, there exists a permutation $\pi$ of members of $A'$ such that $\pi$ is a relaxed plan for $\Pi$.
\end{proof}

\subsection{The diagnostic encoding based on supported models}
One major approach to solving problems in the AI Planning field is to perform backward search, also known as regression, in the search space \cite{planningBook}. In this approach, actions are assumed to act in reverse, i.e., producing their preconditions given they have some effects relevant to the current search node. The main drawback of this approach is that it can easily produce dead-end states, which are not reachable from the initial state. The notion of reversibility of actions has been shown to be quite effective for detecting dead-end states. However, determining the reversibility of actions is itself challenging, and might even need a logic program \cite{FaberMC22} of its own. Nevertheless, the problem of detecting the dead-ends is an easy one in the case of relaxed planning, and can be done in polynomial time as a preprocessing method \cite{HoffmannN01}. Therefore, this backward approach has promise to be efficient for relaxed planning.

Inferring causes from effects can be understood as diagnostic inference \cite{AIbook}. In our causal encoding, we expressed the inference direction from preconditions to dependencies, from dependencies to well-supports, and from well-supports to effects. We can alternatively reverse all these directions to produce a diagnostic encoding. 

In our diagnostic encoding $P_d$, we assume that all atoms could possibly be in the model by using the rule $\set{p}$ for every $p\in X$. However, if $p$ is in the model, then it must  have well-support by at least one action. We establish this by adding $\set{\at{ws}(a,p)}\IF p$ for every $\vec{a}\in A$ such that $p\in add(\vec{a})$, and also $f\IF p, \rg{\naf\at{ws}(a_1,p)}{\AND}{\naf\at{ws}(a_m,p)}, \naf f$ for $p\in X$ and all actions $\vec{a}_1,...,\vec{a}_m$ that could add $p$. The first rule provides the possibility of well-support atoms being in a supported model, while the second rule requires at least one of the well-support atoms to be in the model. To represent the inference from well-supports to dependencies, we add $\at{dep}(p,q)\IF \at{ws}(a,p)$ for $\vec{a}\in A$, $q\in pre(\vec{a})$, and $p\in add(\vec{a})$. Finally, to establish the inference direction from dependencies to preconditions, we add $q \IF\at{dep}(p,q)$. As in $P_c$, all rules in the forms of (\ref{rule:acyc1}) and (\ref{rule:acyc2}) must be included to enforce acyclicity in the supported model. Moreover, we add $a \IF \at{ws}(a,p)$ for $\vec{a}\in A$ and $p\in add(\vec{a})$, to enable an action atom $a$ to represent its cost in the minimization constraint, and also $g\IF\naf g$ for every $g\in G$ to guarantee that goal atoms are included in the model.

It is quite easy to check that if $P_d$ has a supported model $M$, then $M$ is also a supported model of $P_c$. On the other hand, it can be shown in a straightforward manner that if $N$ is a supported model of $P_c$, then $N\setminus L$ is a supported model of $P_d$, where $L$ is the set of atoms $\at{dep}(p,q)$ for which there is no action $\vec{a}$ such that $\at{ws}(a,p)\in N$ and $q\in pre(\vec{a})$. Thus, we have the following result:

\begin{theorem}
\label{prop:diag}
Let $A'$ be any subset of $A^+$. The program $P_d$ has a supported model $M$ such that $A' = \set{\vec{a}\mid a\in M}$ iff $P_c$ has a supported model $N$ such that $A' = \set{\vec{a}\mid a\in N}$.
\end{theorem}

\noindent  Theorem \ref{theorem:causal} and Theorem \ref{prop:diag} can be used to establish Corollary \ref{theorem:diag}.

\begin{corollary}
\label{theorem:diag}
Let $A'$ be any subset of $A^+$. There exists a permutation $\pi$ of members of $A'$ such that $\pi$ is a relaxed plan for $\Pi$ iff $P_d$ has a supported model $M$ such that $A' = \set{\vec{a}\in A^+ \mid a\in M}$.
\end{corollary}

\section{Empirical results}
\label{section:experiments} 

We have implemented our encoding methods inside the HSP* planner~\cite{hsp}. The implementation is available under the ASPTOOLS collection\footnote{https://github.com/asptools/software}. All experiments have been run on a cluster of Linux machines with Intel Xeon 2.40 GHz CPUs, using a timeout of 1800 seconds per problem, and a memory limit of 8 GB. For our supported model based encodings, where vertex elimination is used, for determining the order of vertex elimination, we have implemented the \emph{minimum degree} heuristic, i.e., eliminating a vertex with minimal total number of incoming and outgoing arcs in the graph produced after the elimination of previously eliminated vertices.

Our three implemented encodings are (1) our stable model based encoding $P$; (2) our causal supported model based encoding $P_c$; and (3) our diagnostic supported model based encoding $P_d$. As the solver we use \system{Clasp}~3.3.5, which is capable of optimizing over both stable  and supported models. The \system{Clasp} solver searches for stable models by default. We enable the search for supported models only for our $P_c$ and $P_d$ encodings. As the optimization strategy we use the unsatisfiable core (USC) based search, which our preliminary experiments showed to significantly outperform the branch-and-bound (BB) strategy for the mentioned encodings. Although \system{Clasp} offers a variety of search strategies, we only use the default one. Therefore, the solver parameters have not been tuned to produce the best performance for our new methods. Henceforth, we refer to the method obtained by combing \system{Clasp} with our $P$, $P_c$, and $P_d$ encodings simply by the name of the corresponding encoding. 

To evaluate the efficiency of our methods, we have compared them based on the total time of encoding and solving with 
IP, the integer programming based encoding by \citeN{RankoohRintanen22ICAPS}, which uses IBM ILOG CPLEX Optimization Studio 20.1\footnote{https://www.ibm.com/products/ilog-cplex-optimization-studio} as the optimizer. Regardless of the given time limit, IP has shown to outperform previously introduced methods for optimal relaxed planning including the Boolean satisfiability based encoding used by~\citeN{RankoohRintanen22DR}, the integer programming based model introduced by \citeN{DBLP:journals/jair/ImaiF15}, and the minimum-cost hitting set based method introduced by \citeN{DBLP:conf/aips/HaslumST12}. Since IP has also been implemented inside the HSP* planner~\cite{hsp}, all competing methods share the same code for reading the input problem, grounding, and preprocessing. 

As benchmark problem sets, we use the STRIPS planning problem sets found in the \emph{planning repository}\footnote{https://github.com/AI-Planning/classical-domains}. From IPC domains, domains from both optimal and so-called \emph{satisficing} tracks have been considered. In total, 2212 problem instances from 84 problem sets are used for comparison. Note that this is exactly the benchmark set used in \citeN{RankoohRintanen22ICAPS} for comparing IP with previously introduced methods.

\begin{figure}

    \includegraphics[width=\linewidth]{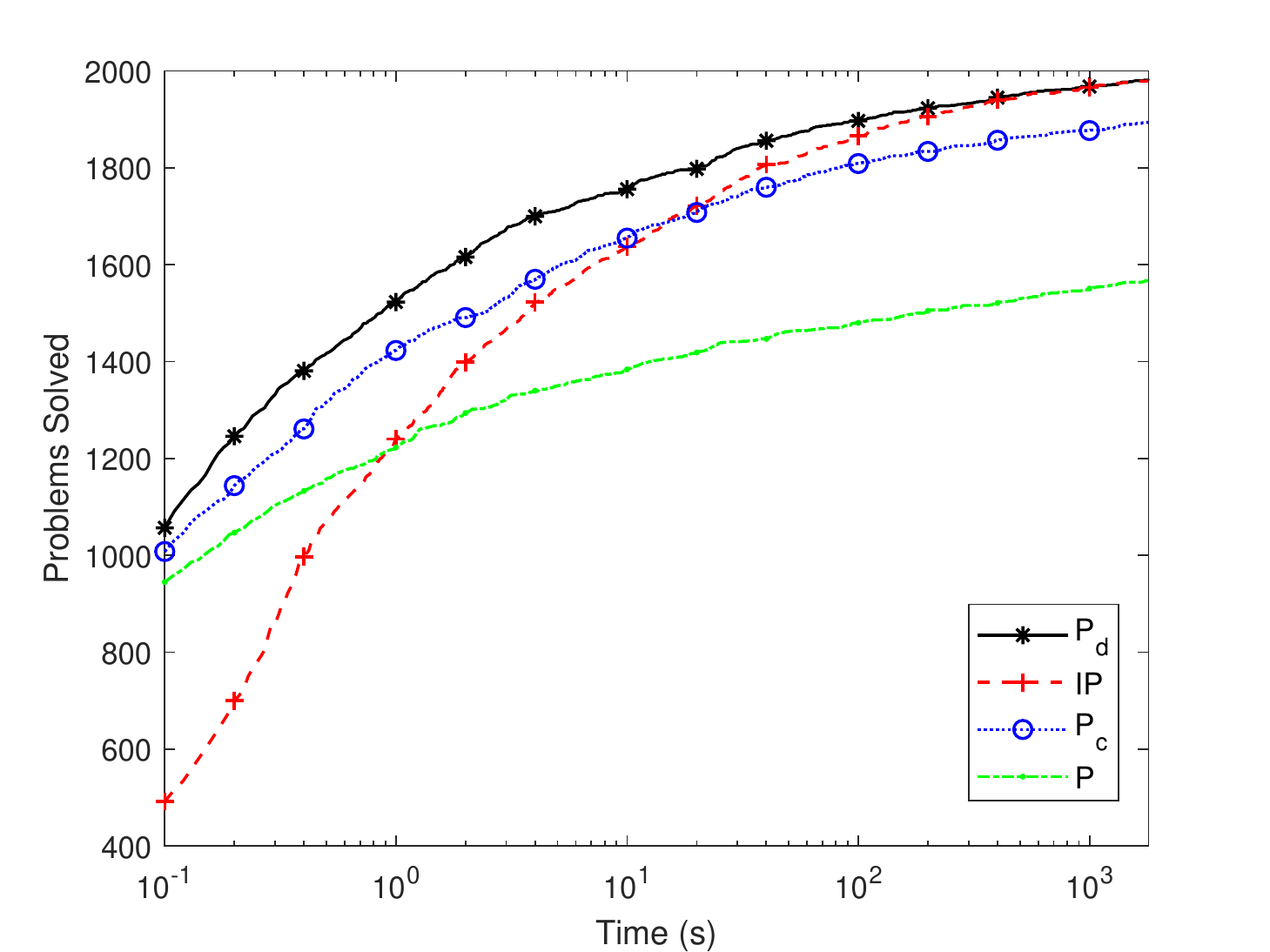}
    \caption{Cumulative numbers of problems solved by the competing methods}
    \label{fig:cactus}
\end{figure}

The cumulative number of problems solved by all methods are presented in Figure \ref{fig:cactus}. Out of the 2212 problems under evaluation, the cost of an optimal relaxed plan was computed in 1800 seconds for 1980, 1982, 1894, and  1567 problems by IP, $P_d$, $P_c$, and $P$, respectively. As it can be seen in Figure \ref{fig:cactus}, our supported model based encodings  significantly outperform the stable model based one, with the diagnostic encoding performing visibly faster than the causal one. Also, even though the number of problems solved within 1800 seconds by our diagnostic encoding is not much higher than that of IP, $P_d$ solves problems considerably faster than IP. In fact, regardless of the time limit, $P_d$ solves more problems compared to any other solver. Particularly, $P_d$ solves 1091 problems in less than 0.1 seconds, more than double the 516 problems solved by IP within the same time limit.

\section{Conclusions and future research}
In this work, we study the previously uninvestigated application of ASP solvers to optimal relaxed planning. Three different encodings of relaxed planning problems into logic programs are provided, one based on the stable model semantics, and two based on the supported model semantics of logic programs. According to our empirical results, all our encodings enable  \system{Clasp} to outperform the state-of-the-art method if the time limit is small. Moreover, our diagnostic supported model based method outperforms the state-of-the-art solver on the studied benchmark problems regardless of the used time limit.

One direction to extend the current work is to study the impact of our new encodings and ASP solvers when employed for computing heuristic values inside state-of-the-art planners. Since our best encoding enables \system{Clasp} to solve almost half of the studied benchmark problems in less than one tenth of second, a direct usage of $h^+$ computed by \system{Clasp} seems to be promising. Also, the usage of USC as the optimization strategy allows for computing lower bounds for $h^+$ within any given time limit. It seems interesting to study the informativeness of such lower bounds in comparison to other commonly used heuristics such as LM-cut, another lower bound of $h^+$, when given the same amount of time for computation.

\label{section:conclusion}

\paragraph{Acknowledgments.}
Financial support from the Academy of Finland
(Project XAILOG, \#345633) is gratefully acknowledged.

\paragraph{Conflict of interest.}
The authors declare no competing interests.


\label{lastpage}
\end{document}